\documentclass[12pt]{iopart}
\usepackage[
backend=biber,
style=alphabetic,
sorting=ynt
]{biblatex}
\newtheorem{definition}{Definition}[section]

\newcommand{\Xc}{\mathcal{X}}
\newcommand{\Yc}{\mathcal{Y}}
\newcommand{\xtrain}{X_{T}}
\newcommand{\ytrain}{Y_{T}}

\DeclareMathAlphabet{\mathsfit}{\encodingdefault}{\sfdefault}{m}{sl}
\SetMathAlphabet{\mathsfit}{bold}{\encodingdefault}{\sfdefault}{bx}{n}

\newcommand{\Ls}{\mathcal{L}}

\usepackage{amsmath,bm}

\usepackage{xargs}                      
\usepackage[colorinlistoftodos,prependcaption]{todonotes}
\newcommandx{\red}[2][1=]{\todo[linecolor=red,backgroundcolor=red!25,bordercolor=red,#1]{#2}}
\newcommandx{\blue}[2][1=]{\todo[linecolor=blue,backgroundcolor=blue!25,bordercolor=blue,#1]{#2}}
\newcommandx{\green}[2][1=]{\todo[linecolor=OliveGreen,backgroundcolor=OliveGreen!25,bordercolor=OliveGreen,#1]{#2}}
\newcommandx{\purple}[2][1=]{\todo[linecolor=Plum,backgroundcolor=Plum!25,bordercolor=Plum,#1]{#2}}

\usepackage{mathrsfs}

\newcommandx{\del}[1]{\dfrac{\mathrm{d} }{\mathrm{d} #1}}

\newcommand{\lp}{\left(}
\newcommand{\rp}{\right)}
\newcommand{\lb}{\left[}
\newcommand{\rb}{\right]}

\newcommand{\linnerprod}{\left\langle}
\newcommand{\rinnerprod}{\right\rangle}

\usepackage{iopams}  
\begin{document}

\begin{center}{\Large \textbf{
Kernels, Data \& Physics\\
}}\end{center}

\begin{center}
Francesco Cagnetta
\textsuperscript{1},
Deborah Oliveira\textsuperscript{2}, 
Mahalakshmi Sabanayagam
\textsuperscript{3},
Nikolaos Tsilivis\textsuperscript{4},
and Julia Kempe\textsuperscript{4}
\end{center}

\begin{center}
{\bf 1} \'Ecole Polytechnique F\'ed\'erale de Lausanne (EPFL)
\\
{\bf 2} Instituto de Matemática Pura e Aplicada (IMPA)
\\
{\bf 3} Technical University of Munich
\\
{\bf 4} New York University
\end{center}

\begin{center}
\today
\end{center}


\section*{Abstract}
{\bf
Lecture notes from the course given by Professor Julia Kempe at the summer school ``Statistical physics of Machine Learning" in Les Houches. The notes discuss the so-called NTK approach to problems in machine learning, which consists of gaining an understanding of generally unsolvable problems by finding a tractable kernel formulation.
The notes are mainly focused on practical applications such as data distillation and adversarial robustness, examples of inductive bias are also discussed.}



\section{Introduction}
\label{sec:introduction}

What exactly makes deep learning work? To answer this question we must recall that deep learning consists of the following three major pillars: the data, the model used to fit the data, and the algorithm used to train the model. The data is characterized by properties such as structure, dimensionality, invariance, and provenance (e.g. if data comes from a physics experiment it must obey certain physical laws); models can be parameterized in different ways depending on the architecture of the network, e.g. the network can have arbitrary depth, its layers can be fully-connected or convolutional; there are several available training algorithms including gradient descent, stochastic gradient descent, Adam, all of which can include regularization of the parameters and potentially lead to a different trained model. The main challenge is then to understand each of these three pillars separately and, more importantly, the connection between them. Moreover, we can address this question from a more theoretical or practical point of view.

Although it is extremely relevant nowadays to understand the big theoretical issues of deep learning, it is also of great importance to pay attention to more practical problems. In fact, there are still major open problems of practical nature, such as adversarial robustness, that we need to solve. In these notes, we are going to address problems of this kind through the lens of NTK methods. We focus in particular on the relationship between data and model, while emphasizing the practical side of the ideas discussed. In that sense, we are going to include the following topics: inductive bias, sample/computational complexity, data distillation, and adversarial robustness.

\subsection{Interplay between theory and practice}

There has been a significant development in the theory of machine learning thanks to the discovery of limiting regimes where the extremely complex learning dynamics of neural networks actually simplify and become analytically tractable. An example is the introduction of the Neural Tangent Kernel (NTK): a kernel method to which neural networks converge in the infinite width limit.
Whenever a new tool such as the NTK is introduced we can ask: are there questions that we could solve with this tool? What are some of the open problems that could benefit from these new lenses? In particular for the NTK, due to its relation with the infinite-width limit: what is the role of overparametrization and how do we explain/harness some of its manifestations, e.g. the phenomenon of double descent and the emergence of an implicit bias? More `practical' questions about learning itself also arise: how do neural networks learn? What do they learn first/last? How fast does learning happen? What properties of the data give rise to successful learning in practice? How are the first phases of learning different from the rest?

\begin{figure}
    \centering
    \includegraphics[width=\textwidth]{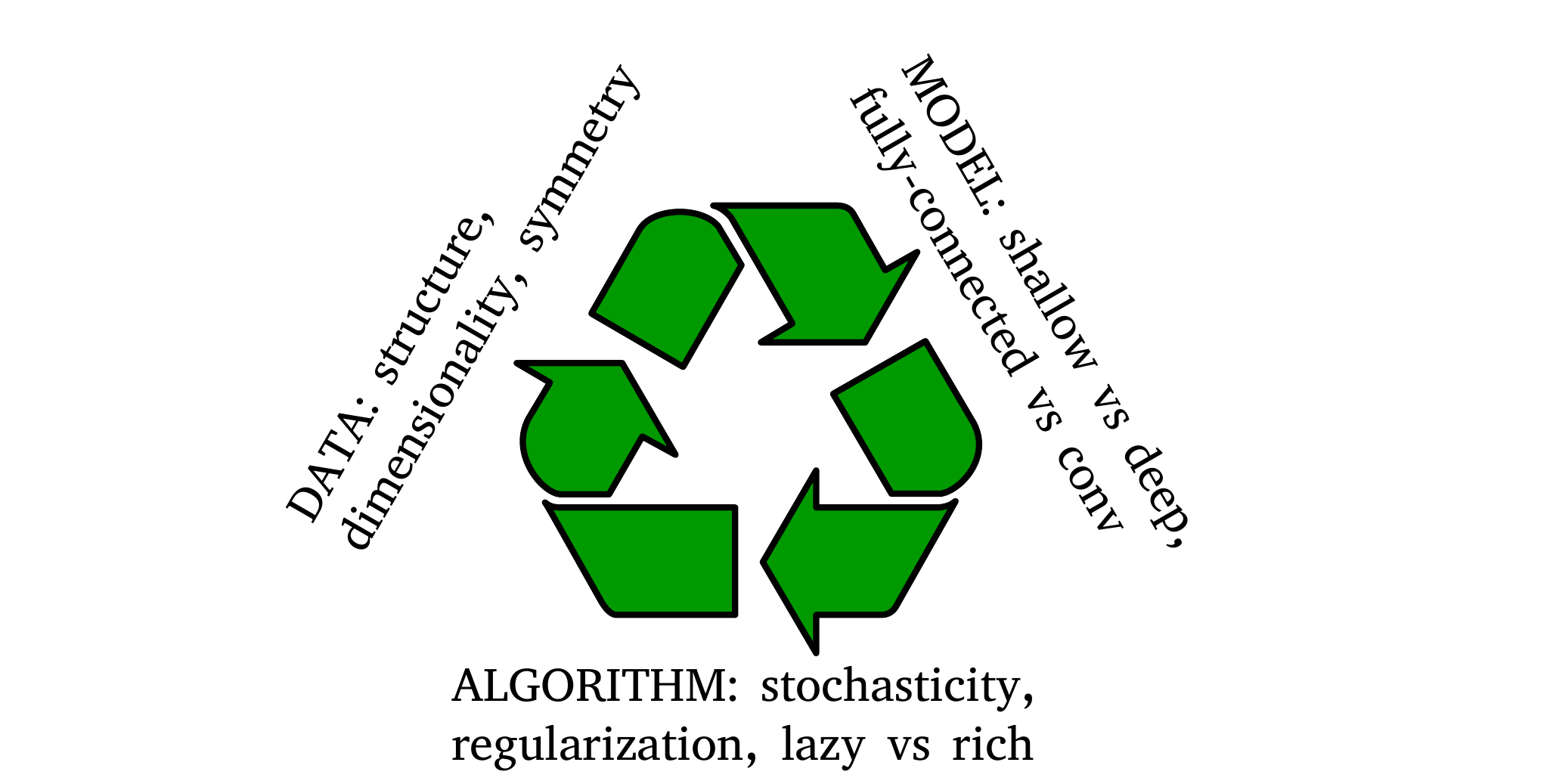}
    \caption{ Pictorial representation of the three pillars of deep learning: data, model, and learning algorithm. Courtesy of Lenka Zdeborov\'{a}.}
    \label{fig:theory-triangle}
\end{figure}

Additionally, we can also ask some questions which are relevant for practitioners: can we create new algorithms (for neural networks) using insights from NTK? Or at least inspired by NTK since it provides us a tractable closed-form expression? Regarding efficient learning, we can pose the question: can we reduce complexity either pre- or post-training? For instance using model distillation, dataset distillation, few-shot learning, and pruning of networks? The aim of this lecture is to understand some of these practical problems. We try to identify where NTK-type methods can be applied, and then try to refine our understanding of deep learning through the solution of these practical problems.

We focus on NTK-based methods for the following reasons: kernel ridge regression is simple, we have a convenient analytical closed-form expression for its training dynamics, and we can take the derivative with respect to the data. In addition, NTKs describe the infinite width limit of neural networks so certain insights/algorithms/techniques may transfer, and, in addition, efficient libraries now exist. A common NTK approach to `practical' problems is the following: 
\begin{enumerate}
    \item Start with a problem for neural networks (intractable, hard, or impossible to solve);
    \item find an underlying NTK formulation;
    \item solve for the NTK setting;
    \item transfer to the neural network setting and hope it works (it does work in the following cases: data distillation, poisoning attacks, pruning, and neural architecture search).
\end{enumerate}
There are still other important problems such as the ones about catastrophic forgetting and continual learning which are not going to be covered in these notes. Important questions about these two problems are: why do neural networks "forget" during continual learning? What methods work to prevent this? How can we create non-forgetting neural networks? And  there is still more that NTKs can do, such as matrix completion, learning "small data" tasks, and recommendation networks. We refer the interested reader, respectively, to  \cite{radhakrishnan2022simple}, \cite{arora2020harnessing} and \cite{sachdeva2022infinite}.

In an attempt to understand some aspects of the relation between the data and the model from a practical perspective, we begin these notes with the topic of inductive bias and its relation with sample/computational complexity.

\subsection{Inductive bias and sample/computational complexity}

Inductive bias refers to the perceived phenomenon that certain architectures are more well-suited to a specific type of data. For example, for many tasks, especially in vision, convolutional (CONV) architectures perform significantly better than their fully-connected (FC) counterparts (at least given the same amount of data). In this case, we say that CONV has a better inductive bias to explain intuitively that CONV matches the underlying structure of the data better, thus models with fewer numbers of parameters generalize better.

How to make the intuitive concept of inductive bias more rigorous? We can try to define a rigorous inductive bias by studying tasks that are easy in some sense for a certain architecture and hard for another. In that way, we hope to find similarities between the properties of these tasks and come up with a useful and rigorous definition. Below we show two tasks that are easily solved by a convolutional neural network but that are hard for a fully-connected one. The tasks presented show different flavors of the inductive bias of CONV: the first one is about sample complexity and the second is about computational complexity.

\paragraph{Sample complexity separation between CONV and FC.} Some tasks require far more training samples on FC than on CONV in order to reach the same test accuracy. This might appear counterintuitive as a large enough FC can simulate a CONV (just set the weight to zero for the edges which are in FC but not CONV),  but the difference in efficiency between the two architectures does not depend on the expressivity of the architecture but on a combination of training algorithm and architecture. Below we give an intuitive explanation, based on the orthogonal equivariance of GD for FC, of why the binary task in $d$ dimensions from \cite{li2020convolutional} is such that CONV needs $\mathcal{O}(1)$ samples but FC needs $\Omega(d)$ samples.

Consider the following setup. Let $x^{i} \in \mathbb{R}^{d}$ and $y^{i} \in \mathbb{R}$ denote the elements of dataset for $i\,{=}\,1,\dots, n$, more compactly $X \in \mathbb{R}^{d \times n}$ and $Y \in \mathbb{R}^{n}$, and let $w \in \mathbb{R}^{d}$ be the vector of weights. We consider linear regression with the squared loss function
\begin{equation}
    \mathcal{L}(w) = \frac{1}{2} \lVert X^{T}w - Y \rVert_{2}^{2}.
\end{equation}
Let $\eta$ be the step size of GD, the iterates are given by
\begin{equation}
    w_{t+1} = w_{t} - \eta \frac{\partial \mathcal{L}}{\partial w} = w_{t} - \eta X(X^{T}w-Y) =: F(w_{t},X,Y),
\end{equation}
and converge to some solution $w^{*} = F^{\infty}(w_{0},X,Y)$. If we change the basis of the data via a rotation matrix $U$, $UX=X'$, we observe that GD applied to the rotated data converges to the rotated solution $w'^{*}=F^{\infty}(w_{0},X',Y)=Uw^{*}$. In fact, we can do a proof by induction with initialization with gaussian weights. As the gaussian distribution is rotationally invariant, $w'_{0}=Uw_{0} = w_{0} \sim \mathcal{N}(0,\sigma^{2}\mathbb{1})$, thus the first step of the induction is verified. Using the induction hypothesis that $w_{t}'=Uw_{t}$ we have
\begin{equation}
    \begin{split}
        w_{t+1}' = F(w_{t}',UX,Y) = w_{t}' - \eta UX(X^{T}U^{T}w_{t}' - Y)\\
        = U\{w_{t} - \eta X(X^{T}w_{t}-Y)\}\\
        = UF(w_{t},X,Y) = Uw_{t+1},
    \end{split}
\end{equation}
as we wished, so that $F$ is \emph{equivariant} under orthogonal transformation. Thus GD is orthogonally equivariant for linear regression, and in fact for any FC architecture: one can prove it by applying the previous argument to each neuron in the first hidden layer of a deep network. 

Due to the equivariance of GD, the result of the algorithm $\mathcal{A}_{FC}(x^1,\dots,x^n)(x)$ is invariant, in the sense that it remains unchanged if the training points $x^i$ and the test point $x$ are rotated. In other words, the algorithm can only argue about the inner products between the data, such that $\mathcal{A}_{FC}(x^{1},...,x^{n})(x) \approx f(\langle x,x^{1} \rangle,\langle x,x^{2} \rangle,...,\langle x,x^{n} \rangle)$. To see how this affects sample complexity, consider the dataset $\{x^i=e_{i}y^i,y^i\}$ where $y^i \in \{\pm 1\}$ and the $e_{i}\in \mathbb{R}^{d}$ are the canonical basis vectors. If we assume $n<d/2$, then we have $\langle x,x^{j} \rangle = 0$ for all $j$'s with probability greater than $1/2$, i.e., the new point lies outside the subspace generated by the previous ones with probability $>1/2$. In this case, the best thing to do is to guess the sign, thus we get a wrong prediction with probability $1/2$. This implies that $\mathcal{A}_{FC}$ is wrong with probability greater than $1/4$, thus we need at least $d/2$ samples for FC to perform well.

In~\cite{li2020convolutional} they actually have $x \sim \mathcal{N}(0,\mathbb{1}_{2d})$ and the labels are
\begin{equation}
    f(x) = \mathrm{sign}\left(\sum_{i=1}^{d}x_{i}^2-\sum_{i=d+1}^{2d}x_{i}^2\right).
\end{equation}
It is relatively easy to construct a CONV network that can learn this function $f$. Roughly speaking, now that we are dealing not only with the unit cube but with the whole space, in order to choose a subspace (i.e. distinguish between the two outputs of f) we can use rotation matrices that rotate only that subspace, requiring $\Omega(d^2)$ samples.

\paragraph{Computational complexity separation between CONV and FC.} The second task we examine can be efficiently solved by CONV while provably hard for FC with gradient descent~\cite{shalev2020computational}. In such a task data have a hidden consecutive pattern, so that polynomial size CONV needs $Poly(d)$ steps but polynomial size FC needs $Superpoly(d\log{d})$ steps to identify it.

More explicitly the task is the following: we have a function $f:\{\pm 1\}^{d} \rightarrow \{\pm 1\}$ and a consecutive hidden pattern, i.e. $\exists j^{*}$ and a function $g:\{\pm 1\}^k \rightarrow \{\pm 1\}$ such that \linebreak $f(x_{1},\dots,x_{d}) = g(x_{j^{*}},\dots,x_{j^{*}+k-1})$, where $k \sim \log{d} \ll d$; the aim is to discover $g$ and the hidden sequence. For the purpose of identifying the consecutive pattern, the algorithm needs to be able to distinguish between permutations of the $d$-dimensional input, but GD on FC is equivariant under permutations. Thus, for the same argument used in the previous section, this task is hard for FC. Tasks of this kind, which can be represented as a Boolean function depending on only $k$ out of $d$ variables, are called \emph{$k$-juntas} and are notoriously difficult to solve.  

Consider however the following two-layer CONV neural network,
\begin{equation}
    f_{u,w,b}(x) = \sum_{j=1}^{n-k} \langle u^{j}, \sigma(Wx_{j}...x_{j+k-1} +b)\rangle
\end{equation}
where $u^{j} \in \mathbb{R}^{q}$ is the outer layer, $W \in \mathbb{R}^{q\times k}$ the inner layer weights, $b\in \mathbb{R}^{q}$ the bias and $x_{j}...x_{j+k-1}$ denotes a sequence of $k$ input bits beginning at $j$. We assume that $W^{(0)}\sim\{\pm 1/k\}^{q\times k}$, $b_{i}=1/k-1$ at initialization, and the number of filters is $q > 2^{k+3}\log{(2^{k}/\delta)}$ for some $\delta>0$.

Finding the hidden sequence is easy for a two-layer CONV network with $q$ channels and windows of size $k$ (in the sense that there exists a GD algorithm that performs well with a relatively small number of steps) because of the following intuition. The number of channels $q$ required is related to the coupon collector problem: if there are $n$ different coupons and you want to collect them all, you need on average to draw with replacement $n\log{n}$ coupons. So we need on average $2^{k}\log{2^{k}}$ channels to learn the values of $g$ on all the possible inputs ($2^k$), i.e. to learn $g$. In fact, the choices of weights and biases cause the output to lie in the set $\{-2+1/k, ..., -1/k, 1/k \}$ of $2^k$ distinct elements with just one positive, such that an activation function like ReLU can single out the only positive element. This means that each filter, i.e., each line of the weight matrix, is an indicator function for exactly one value of the input $x$. We use a coupon-collector number of them to ensure we can learn $g$ with certainty because we can always write $g$ as a combination of these indicators:
\begin{equation}
    g(x) = \sum_{z\in \{\pm 1\}^k} \delta_{zx}g(z)
\end{equation}
where $\delta_{zx}$ is the Kronecker delta, $z_{i}=\mathrm{sign}(W_{z})_{i}$ and $W_{z} \in \{\pm 1/k\}^{k}$ is the filter representing the input $z \in \{\pm 1\}^{k}$. The delta function can be achieved by a filter of weights of size $k$. As we have $2^k$ possible different inputs, and the weights are initialized at random we need (coupon collector) $2^{k}\log{2^k}$ filters to distinguish the output of the function with high certainty. As we are assuming $k \sim \log{d}$ we have $2^k \sim d$ and this means that we only need to do linear interpolation in the next layer to learn the function, in other words, a relatively easy problem using gradient descent.

\section{The Neural Tangent Kernel perspective}
\label{sec:ntk}

The Neural Tangent Kernel was first introduced in~\cite{jacot2018neural} as the object which captures the dynamics of artificial neural networks in function space in the infinite-width limit. More specifically, we consider an \emph{artificial neural network} as a mapping from a set of $P$ parameters $\theta \in \mathbb{R}^P$ to a function $f_\theta:\Xc\to \mathbb{R}$. For any input $x\in\mathcal{X}$, denote with $f_\theta(x)$ the value of $f_\theta$ in $\bm{x}$. The definition of the neural tangent kernel follows~\cite{jacot2018neural}.
\begin{definition}[Neural Tangent Kernel (NTK)]
Consider an artificial neural network $f$ with parameters $\theta\in\mathbb{R}^P$ such that $f_\theta:\mathbb{R}^d\to \mathbb{R}$, let $\theta_p$ denote the $p$-th component of $\theta$ for any $p\,{=}\,1,\dots,P$. The Neural Tangent Kernel $\mathcal{K}_\theta$ of the network is a function on $\mathbb{R}^d\times \mathbb{R}^d$ defined as
\begin{equation}\label{eq:NTK}
    \mathcal{K}_\theta(x,x') := \sum_{p=1}^P \partial_{\theta_p} f_\theta(x)\partial_{\theta_p} f_\theta(x') = \left(\nabla_\theta f_\theta(x)\right)^T \nabla_\theta f_\theta(x'),
\end{equation}
where we have introduced $\nabla_\theta$ as the gradient w.r.t. $\theta$.
\end{definition}
It turns out that when \emph{i)} $f_\theta$ is parametrized as a classic feed-forward neural network and \emph{ii)} with proper initialization of the parameters $\theta_{\text{init}}$ \emph{then}, in the limit where all hidden layers of the network have infinite width, the initial NTK $\mathcal{K}_{\theta_{\text{init}}}$ converges to a parameter-independent limit $\mathcal{K}$ and remains constant while the parameters are updated via gradient descent~\cite{jacot2018neural}.

\subsection{Motivation \& Notation}

Where does the definition of the NTK come from? It emerges naturally by considering the dynamics of the network in function space, as we show in the following.

\paragraph{Notation and general setup.} Let $\Xc$ denote some input space, $\Xc\subset \mathbb{R}^d$ and $\Yc$ some output space, $\Yc\subset \mathbb{R}$. Let us define the \emph{training set} as a set of $n$ input-output pairs $(x_i,y_i)_{i=1,\dots,n} \in \left(\mathcal{X}\times\mathcal{Y}\right)^n$. The goal of supervised learning is that of finding, given a training set, some parameters $\theta^*$ such that $f_{\theta^*}:\mathbb{R}^d\to \mathbb{R}$ approximates the functional relationship between inputs and outputs in the training set. For instance $\theta^*$ can be found via \emph{empirical risk minimization}: fix a \emph{loss function} $\ell:\mathbb{R}\times\mathbb{R}\to\mathbb{R}$ (e.g. $\ell(f,y)\,{=}\,(f-y)^2/2$) and choose $\theta^*$ as the minimizer of the empirical average of the loss over the training set,
\begin{equation}\label{eq:empirical-risk-min}
    \theta^* = \mathrm{arg} \min_\theta\left\lbrace\displaystyle \sum_{i=1}^n \ell\left(f(x_i;\theta),y_i\right)\right\rbrace.
\end{equation}
In practice, the parameters $\theta$ are set to some initial value $\theta_{\text{init}}$, then updated by descending along the gradients $\nabla_\theta$ of the empirical loss function at some learning rate $\eta$. Here we consider a continuous-time \emph{Gradient-Flow} (GF) dynamics,
\begin{align}\label{eq:gd-cont-loose}
    \frac{d}{dt}\theta_t &= -\eta\sum_{i=1}^n \left.\nabla_\theta \ell\left(f(x_i;\theta),y_i\right)\right|_{\theta=\theta_t},\quad \theta_0=\theta_{\text{init}},
\end{align}
which yields the usual Gradient Descent (GD) dynamics upon discretizing the time-derivative on the left-hand side with step size $ d t\,{=}\,1$. 

\paragraph{Gradient-flow in function space.} By differentiating $f(\theta_t)$ w.r.t. $t$ and applying the chain rule we immediately get an equation for the network's dynamics in function space:
\begin{align}\label{eq:gd-func-loose}
    \frac{d}{dt}f(x;\theta_t) = &\left(\left.\nabla_{\theta}
f_\theta(x)\right|_{\theta=\theta_t}\right)^T\frac{d}{dt}\theta_t \nonumber \\
= &\left(\left.\nabla_{\theta}
f_\theta(x)\right|_{\theta=\theta_t}\right)^T \left( -\eta\sum_{i=1}^n \left.\nabla_{\theta} \ell\left(f_\theta(x_i),y_i\right)\right|_{\theta=\theta_t} \right) \nonumber \\
= &-\eta\sum_{i=1}^n\left(\left.\nabla_{\theta}
f_\theta(x)\right|_{\theta=\theta_t}\right)^T\left( \left.\partial_f \ell\left(f,y_i\right)\right|_{f=f_{\theta_t}(x_i)} \left.\nabla_{\theta} f_\theta(x_i)\right|_{\theta=\theta_t} \right) \nonumber \\
= &-\eta\sum_{i=1}^n \left(\left.\left(\nabla_\theta f_\theta(x)\right)^T \nabla_\theta f_\theta(x_i)\right|_{\theta=\theta_t}\right)\left.\partial_f \ell\left(f,y_i\right)\right|_{f=f_{\theta_t}(x_i)},
\end{align}
where we recognize the NTK $\mathcal{K}_{\theta_t}(x,x_i)$ between round brackets. For any training set of size $n$ $(x_i,y_i)_{i=1,\dots,n}$ let us denote with $\xtrain$ and $\ytrain$ the $n$-dimensional column vectors obtained by stacking the $x_i$'s and the $y_i$'s, respectively. Analogously, denote with $g(\xtrain, \ytrain)$ the vector obtained by element-wise application of any function $g:\Xc\times\Yc\to\mathbb{R}$ to the training set, so that ~\eqref{eq:gd-func-loose} can be written compactly as,
\begin{align}\label{eq:gd-func}
    \frac{d}{dt}f_{\theta_t}(x) = -\eta \left.\mathcal{K}_{\theta}(x,\xtrain) \cdot \partial_f \ell\left(f_\theta\left(\xtrain\right),\ytrain\right) \right|_{\theta=\theta_t}.
\end{align}
Notice that, if $\mathcal{K}_{\theta_t}$ converges to a parameter-independent limit for all $t$, then~\eqref{eq:gd-func} does not depend explicitly on the parameters $\theta$.

\subsection{The infinite-width limit}

\eqref{eq:gd-func} elucidates the general relationship between the NTK and the dynamics of any network in function space. As anticipated, further simplifications occur under additional conditions on the network, as the NTK converges to a time-independent and parameter-independent limit $\mathcal{K}(x,x')$~\cite{jacot2018neural}. To illustrate this convergence we follow the approach of~\cite{lee2019wide} and first consider the dynamics of the network when the parameters remain arbitrarily close to their initial values, then show that this is indeed the case when the width of all the network layers is very large.

\paragraph{The NTK parametrization.} More specifically, consider an artificial neural network of depth $L\,{+}\,1$ with the following parametrization. With $d_0\,{=}\,d$ (the dimension of input space $\Xc$), $W^{(l)}$ a $d_{l}\times d_{l-1}$ matrix for all $l\,{=}\,1,\dots,L$, $W^{(L+1)}$ a $d_L$-dimensional row vector,
\begin{equation}\label{eq:deepNN}
    f_\theta(x) = W^{(L+1)}\sqrt{\frac{c}{d_L}} \sigma\left( W^{(L)}\sqrt{\frac{c}{d_{L-1}}} \sigma\left( \dots \sigma\left(\frac{W^{(1)}x}{\sqrt{d_0}}\right)\dots\right) \right),
\end{equation}
where $\sigma:\mathbb{R}\to\mathbb{R}$ is the activation function and $c$ a $\sigma$-dependent normalization factor such that $\mathbb{E}_{z\sim\mathcal{N}(0,1)}[\sigma(z)^2]\,{=}\,c^{-1}$~\footnote{$\mathbb{E}_{z\sim\mathcal{N}(0,1)}$ denotes expectation with respect to zero-mean, unit-variance Gaussian random variable $z$.}. The initial set of parameters $\theta_{\text{init}}$ is obtained by drawing all the elements of the matrices $W^{(l)}$ independently from a zero-mean, unit-variance Gaussian distribution. Notice that, as the widths of the $L$ hidden layers diverge, e.g. $d_1\,{=}\,\dots\,{=}\,d_L\,{=}\,m\to \infty$, the initial network $f_{\theta_{\text{init}}}(x)$ converges to a Gaussian process~\cite{neal2012bayesian, daniely2016towards}. However, for the sake of simplicity, we will assume that $f_{\theta_{\text{init}}}(x)\,{=}\,0$ for all $x$---this condition can be realized by initializing two identical networks of size $d_L/2$ at the last hidden layer and setting $f_{\theta}(x)$ equal to their difference.

\paragraph{Linearized dynamics.} Let us replace, following~\cite{lee2019wide}, the network function with its first-order Taylor expansion around $\theta_{\text{init}}$, i.e. (since $f(x;\theta_{\text{init}})\,{=}\,0$)
\begin{equation}\label{eq:linearized}
    f^{\text{lin}}_{\theta_t}(x) = \left(\left.\nabla_{\theta} f_{\theta}(x)\right|_{\theta=\theta_{\text{init}}}\right)^T (\theta_t-\theta_{\text{init}}).
\end{equation}
By plugging the linear expansion into~\eqref{eq:gd-func} we get
\begin{equation}\label{eq:gd-func-lin}
    \frac{d}{dt}f^{\text{lin}}_{\theta_t}(x) = -\eta\mathcal{K}_{\theta_{\text{init}}}(x,\xtrain) \cdot \left.\partial_f \ell\left(f^{\text{lin}}_\theta\left(\xtrain\right),\ytrain\right) \right|_{\theta=\theta_t}.
\end{equation}
Since $\mathcal{K}$ does not change during training, these dynamics are much simpler than~\eqref{eq:gd-func}, in that they do not depend explicitly on the parameters $\theta_t$. In particular, if the loss is the square loss $\ell(f,y)\,{=}\,(f-y)^2/2$, then~\eqref{eq:gd-func-lin} becomes a linear equation in $f^{\text{lin}}$ which is easily solved. For the values of $f^{\text{lin}}$ on the training set, for instance,
\begin{align}\label{eq:gd-func-lin-train}
    &\frac{d}{dt}f^{\text{lin}}_t(\xtrain) = -\eta\mathcal{K}(\xtrain,\xtrain) \cdot \left(f^{\text{lin}}_t(\xtrain)-\ytrain\right),\quad f^{\text{lin}}_0(\xtrain)=0,\nonumber\\
    &\Rightarrow f^{\text{lin}}_t(\xtrain) = \left(\mathbb{I}-e^{-\eta\mathcal{K}(\xtrain,\xtrain)t}\right) \cdot\ytrain,
\end{align}
where we have removed the argument $\theta_{\text{init}}$ from $\mathcal{K}$ to ease the notation and $\mathbb{I}$ denotes the $n\times n$ identity matrix. ~\eqref{eq:gd-func-lin-train} shows that linearized neural networks have a simple dynamics in function space which is entirely controlled by the neural tangent kernel at initialization.

\paragraph{Infinite-width limit $\equiv$ linearized dynamics.} It turns out that, when the widths of all the hidden layers of the network are sufficiently large, the parameters $\theta_t$ remain infinitesimally close to their initialization $\theta_{\text{init}}$. As a result, the linearized dynamics become a good approximation of the general dynamics. More specifically, let us set the number of hidden layers $L$ to $1$  and $d_1\,{=}\,m$---the general case with depth $L$ and $d_1\,{=}\,\dots\,{=}\,d_L\,{=}\,m$ is discussed in~\cite{lee2019wide, arora2019exact}. For $L\,{=}\,1$, the parameters consist of the $m$-dim. row vector $W^{(2)}$ and the $m\times d_0$ matrix $W^{(1)}$, made in turn of $m$ $d_0$-dim. row vectors $W^{(1)}_i$. By plugging~\eqref{eq:deepNN} with $L\,{=}\,1$ into the NTK definition~\eqref{eq:NTK} we get
\begin{equation}
\mathcal{K}_{\theta}(x,x') = \frac{c}{m} \sum_{i=1}^m \left[\sigma\left(\frac{W^{(1)}_i x}{\sqrt{d_0}}\right)  \sigma\left(\frac{W^{(1)}_i x'}{\sqrt{d_0}}\right) + (W^{(2)}_i)^2
\sigma'\left(\frac{W^{(1)}_i x}{\sqrt{d_0}}\right)  \sigma'\left(\frac{W^{(1)}_i x'}{\sqrt{d_0}}\right) \frac{x^T x'}{d_0}\right],
\end{equation}
with $\sigma'$ denoting the derivative of the activation function. The $m\to\infty$ limit of the NTK at initialization is given by the law of large numbers~\footnote{ $\mathbb{E}_{w\sim\mathcal{N}_(0,\mathbb{I}_{d_0})}$ denotes expectation with respect to a zero-mean, identity-covariance $d_0$-dim. Gaussian vector.}:
\begin{align}
    &\mathcal{K}_{\theta_{\text{init}}}(x,x') \xrightarrow{m\to\infty}\mathcal{K}(x,x')\quad\text{(with prob. $1$)},\nonumber\\&\mathcal{K}(x,x') := c\mathbb{E}_{w\sim\mathcal{N}_(0,\mathbb{I}_{d_0})} \left[\sigma\left(\frac{w^T x}{\sqrt{d_0}}\right)  \sigma\left(\frac{w^T x'}{\sqrt{d_0}}\right) +
\sigma'\left(\frac{w^T x}{\sqrt{d_0}}\right)  \sigma'\left(\frac{w^T x'}{\sqrt{d_0}}\right) \frac{x^T x'}{d_0}\right].
\end{align}
Consider now the following assumptions:
\begin{itemize}
    \item the matrix $\mathcal{K}(\xtrain, \xtrain)$ is full rank, i.e. $0\,{<}\,\lambda_{\text{min}}(\mathcal{K})\,{<}\,\lambda_{\text{max}}(\mathcal{K})\,{<}\,+\infty$;
    \item the input space $\Xc$ is compact and the training points within a training set $(x_i, y_i)_{i\,{=}\,1,\dots,n}$ are all distinct;
    \item the activation function $\sigma$ is Lipschitz and bounded over the input space ($\| \sigma \|_{\infty}\,{<}\,\infty$).
\end{itemize}
Under these assumptions, building on the local Lipschitzness of the network gradients $\nabla_\theta f_\theta(x)$ around initialization, it is easy to show the following. For any learning rate $\eta\,{<}\,2(\lambda_{\text{min}}+\lambda_{\text{max}})^{-1}$ and $R_0$ bounding the initial training loss with high probability, the following bounds hold with high probability when $m$ is large:
\begin{align}
    \| f_{\theta_t}(\xtrain)-\ytrain \|^2 \leq e^{-\frac{2\eta \lambda_{\text{min}}t}{3}} R_0,\quad \| \theta_t -\theta_{\text{init}}\| \leq \frac{3 K R_0}{\lambda_{\text{min}}} (1-e^{-\frac{\eta \lambda_{\text{min}}t}{3}}) m^{-1/2},\nonumber\\
    \text{sup}_t \| \mathcal{K}_{\theta_t}(\xtrain,\xtrain) -\mathcal{K}(\xtrain,\xtrain) \| \leq\frac{6 K^3 R_0}{\lambda_{\text{min}}} m^{-1/2}.
\end{align}
As a result, the infinite-width dynamics of networks like~\eqref{eq:deepNN} coincide with the linearized dynamics~\eqref{eq:gd-func-lin}.

\subsection{Simple applications of the NTK approach}
\label{ssec:examples}

The linearization of the gradient-descent dynamics entails a profound simplification of any learning problem which admits an NTK limit. In addition, the infinite-time solution of the linearized dynamics~\eqref{eq:gd-func-lin} coincides with the predictor of kernel regression. In particular, since we assumed that the initialized network $f_{\theta_{\text{init}}}$ coincides with the $0$-function, for the square loss one has
\begin{equation}\label{eq:krr-predictor}
    f^{\text{lin}}_{\infty}(x) = \left(\mathcal{K}(x, X_T)\right)^T\left(\mathcal{K}(X_T, X_T)\right)^{-1} Y_T.
\end{equation}
Thus all the tools of the theory of kernel regression can be deployed to study neural networks---above all the closed-form expression for the trained predictor given the training set~\eqref{eq:krr-predictor}. These two aspects justify the validity of the so-called NTK approach as a tool for a theoretical understanding of neural networks, which we repeat here for the sake of completeness.
\begin{enumerate}
    \item Start with a problem for neural networks (intractable, hard or impossible to solve);
    \item find an underlying NTK formulation;
    \item solve for the NTK setting;
    \item transfer the knowledge acquired to the neural network setting (and hope it works).
\end{enumerate}
In this section we focus on two simple examples of the NTK approach: one is the proof of the spectral bias phenomenon (~\eqref{sssec:spectral}), and the other is a state-of-the-art data distillation method based on Kernel-Inducing-Points~\cite{nguyen2021dataset} (~\eqref{sssec:distillation}).

\subsubsection{Spectral bias}
\label{sssec:spectral}

The linearized dynamics ~\eqref{eq:gd-func-lin-train} can be used to deduce how the empirical loss approaches zero over training. After introducing the eigendecomposition of the kernel (eigenvalues $\lambda_i$, eigenvectors $V_i$),
\begin{equation}
    \mathcal{K}(\xtrain,\xtrain) = \sum_{i=1}^n \lambda_i V_i  V_i^T,
\end{equation}
we get
\begin{equation}
\displaystyle \sum_{i=1}^n \left(f_t^{\text{lin}}(x_i)-y_i\right)^2 =  \| e^{-\eta\mathcal{K}(\xtrain,\xtrain)t}\cdot\ytrain \|^2 = \sum_{i=1}^n e^{-2\eta \lambda_i t } \left(V_i \cdot \ytrain\right)^2,
\end{equation}
which shows that the loss converges to zero exponentially fast, at a rate dictated by the learning rate and the smallest eigenvalue of $\mathcal{K}(\xtrain,\xtrain)$. Choosing the maximal learning rate $\eta = 2/\lambda_{\text{max}}$, convergence is controlled by the condition number $\kappa\,{=}\,\lambda_{\text{min}}/\lambda_{\text{max}}$. 

In simple terms, the spectral bias says that a neural network learns progressively `complex' functions during training. The analysis above shows that this is indeed the case in the NTK limit if the `complexity' of a function is measured with the projections onto the kernel eigenfunctions $V_i$ ordered according to the magnitude of the corresponding eigenvalue~\cite{basri2020frequency}. This observation has given rise to a whole line of work focused on explaining the properties of neural networks by computing the NTK spectrum under a given data distribution. For instance, studying such spectrum allows one to sort functions according to how easily they can be learned with a neural network~\cite{bach2017breaking, bietti2019inductive}, or to prove that fully-connected networks of different depths lead to essentially the same machine-learning method in the NTK limit~\cite{bietti2021deep}. In addition, since the spectrum can be related to the generalization capabilities of the network~\cite{bordelon2020spectrum, spigler2020asymptotic, loureiro2021learning, tomasini22failure}, this approach also allows one to study the interplay of architecture, data structure, and generalization by considering the NTK of convolutional architectures, whether shallow~\cite{favero2021locality, bietti2022approximation} or deep~\cite{cagnetta2022wide, xiao2021eigenspace} or graph neural networks~\cite{du2019graph}.

\subsubsection{Data distillation via Kernel-Inducing-Points}
\label{sssec:distillation}

\emph{Data distillation} is a significant reduction in the dataset size which is achieved by creating a small synthetic dataset such that a machine learning algorithm would learn as efficiently as if it were learning on the full data. It builds on the concept of knowledge distillation~\cite{hinton2015distilling} and was first proposed in \cite{Wan+18}.

The concept of data distillation can be understood by thinking of support vector classifiers as an extreme example. Consider specifically the problem of maximum-margin classification of linearly separable data in $d$ dimensions. Given a set of training examples, the maximum-margin classifier is the $(d-1)$-dimensional hyperplane for which the distance with the closest training points is the largest. Once such a hyperplane is identified, its position depends only on the closest training points, or \emph{support vectors}. Therefore, replacing the full training set with the support vectors results in a significant reduction of the dataset size while leaving the predictor unchanged---data distillation is achieved. Support vectors provide a very specific example, in that the elements of the distilled dataset are also elements of the original training set. In general, the distilled data are different from any other datum. The practical relevance of data distillation is obvious since it results in a significant reduction of the dataset size by definition. The theoretical implications are more subtle: what does data distillation tell us about the amount of information encoded in the training set? Can one unveil the properties of natural data by building distilled datasets (e.g. low intrinsic dimensionality~\cite{pope2021intrinsic})?

In quantitative terms, Data Distillation (DD) can be defined as follows. Consider a `support' dataset $X_{S}$, as opposed to the full training set $X_{T}$. The goal of DD is that of learning $X_{\text{S}}$ such that a given machine learning method trained on the support set gives the same result as if trained on the full training set. This is a daunting task in general, as it requires knowing the relationship between the training set and the trained model, which is itself a major problem in the theory of deep learning. In mathematical terms, the problem is formulated as nested minimization. Let us first introduce a small change of notation so as to make the dependence of the loss on the training set and network parameters explicit (following~\cite{Wan+18}),
\begin{equation}
    \ell\left(f_\theta\left(\xtrain\right),\ytrain\right) \to \ell\left(X_T, \theta \right).
\end{equation}
Thus, given $X_S$, we find the parameters by minimizing the loss with training set $X_S$, then find the $X_S$ such that the loss on the full training set $X_T$ is minimal: 
\begin{equation}
    \mathrm{arg} \min_{X_S} \ell\left(X_T, \mathrm{arg} \min_\theta \ell\left(X_S, \theta \right) \right)
\end{equation}

Even simplified formulations of DD, such as the one introduced in~\cite{Wan+18}, entail a number of non-trivial challenges. More specifically, a support set $X_S$ can be found by simply asking that a single step of gradient descent on $X_S$ also decreases the loss on $X_T$ thus bypassing the `inner' minimization problem.
Starting from an arbitrary set of parameters $\theta_0$, a step of gradient descent on the support set $X_S$ with learning rate $\eta$ leads to
\begin{equation}
    \theta_1 = \theta_0 -\eta \left.\nabla_\theta \ell(X_S,\theta)\right|_{\theta=\theta_0}.
\end{equation}
Thus, asking for the couple $\eta$, $X_S$ which yields the largest decrease of the total loss is equivalent to the following problem,
\begin{equation}\label{eq:data-dist}
    \mathrm{arg} \min_{\eta, X_S} \ell\left( X_{\text{T}}; \theta_0-\eta \nabla_{\theta} \ell(X_{\text{S}},\theta)|_{\theta=\theta_0}\right).
\end{equation}
In order to solve~\eqref{eq:data-dist} in practice, one has to first follow the gradient descent step,
\begin{equation}\label{eq:data-dist-1}
    \theta_1 = \theta_0-\eta \nabla_{\theta} \ell(X_{\text{S}},\theta)|_{\theta=\theta_0},
\end{equation}
then update both $X_S$ and $\eta$ with another learning rate $\alpha$, along the gradients of the full training loss $\ell(X_T, \theta_1)$. This method depends heavily on the initialization $\theta_0$, meaning that data distillation would only work on models initialized with $\theta_0$---the same initialization used for learning $X_{\text{S}}$. Although this problem can be solved (e.g. by repeating the gradient descent step for different initializations then averaging over all such initializations) it illustrates the many challenges that emerge from the nested optimization problem required to solve DD.

\begin{figure}
    \centering
    \includegraphics[scale=0.3]{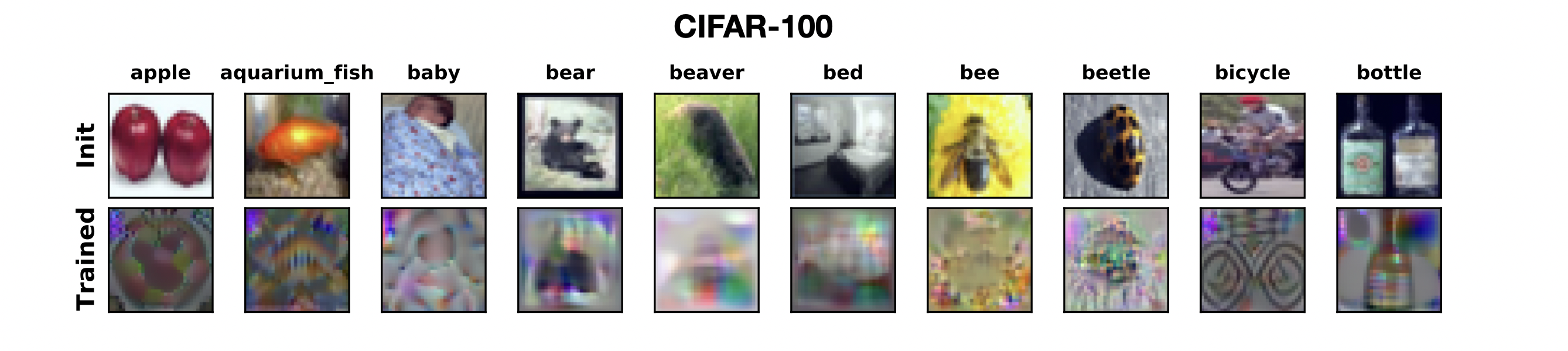}
    \caption{Examples of learned images on the CIFAR-100 benchmark. Images are initialized from natural images in the top row and converge to images in the bottom row using the KIP algorithm \cite{Ngu+21}.}
    \label{fig:kip_img}
\end{figure}

However, in the NTK limit, one can write the value of the predictor trained on the support set $X_S$ in closed analytic form, so that DD simplifies greatly. With the square loss, in particular, using~\eqref{eq:krr-predictor} with $X_S$ as training set and $X_T$ as test set, DD can be reformulated as
\begin{equation}
\mathrm{arg} \min_{X_S} \|\ytrain - \mathcal{K}(\xtrain, X_S)\mathcal{K}(X_S, X_S)^{-1} Y_S\|^2.
\end{equation}
The loss now has an analytic expression as a function of $X_S$ which can be easily differentiated, thus easing the search for an optimal $X_S$ by gradient descent. This method was introduced in~\cite{nguyen2021dataset} and the elements of the support set dubbed Kernel Inducing Points (KIP, see Figure \ref{fig:kip_img} for an example of distilled dataset). Despite the enormous simplification obtained by moving to the NTK limit, using KIP turns out to produce state-of-the-art results for data distillation~\cite{Ngu+21}.

\section{Adversarial Robustness}
Deep learning methods produce state-of-the-art results for several challenging problems despite a lack of understanding. By contrast, adversarially robust systems are still difficult to obtain, even in practice. The major challenge in adversarial robustness is that it requires a formulation as a min-max problem, that is finding a saddle point. This problem is generally more difficult to solve than other deep learning methods which can be cast as minimization problems. Owing to this additional complexity, several aspects of adversarial robustness are not well understood, not even at an intuitive level. 

\paragraph{Adversarial examples.} Adversarial vulnerability of deep learning models occurs when a small perturbation of a datum, e.g. a change in an image that is imperceptible to humans, leads to drastically wrong predictions~\cite{goodfellow2014explaining}. Adversarial attacks have been found to be pervasive \cite{wu2020defense, chen2018shapeshifter} and interestingly these attacks are universally transferable. The idea of adversarial attack is formalized mathematically with \emph{adversarial examples}. Let $x \in \mathbb{R}^d $ be data features, $f(x)$ denote the output of some model on $x$, and denote the loss with $\ell$. With $d(.)$ denoting some $\ell_p$-norm distance, $\tilde{x}$ is adversarial if $d(x,\tilde{x}) \leq \epsilon$ and $f(x) \ne f(\tilde{x})$. Formally,
\begin{equation}
\tilde{x}= \underset{\mathcal{B}_{\epsilon}(x)}{\mathrm{arg} \max}\, \mathcal{\ell}(f(\tilde{x}),y),
\end{equation}
where $\mathcal{B}_{\epsilon}(x)$ denotes the $\ell_p$-norm ball of radius $\epsilon$ around $x$, also referred to as the \emph{budget}.
Initially~\cite{goodfellow2014explaining} the presence of adversarial examples was explained as a high-dimensional phenomenon with the following reasoning: Let $f(x)=w^Tx$, for the adversarial example $\tilde{x}=x + \delta$, $f(\tilde{x})=w^Tx + w^T\delta$ where $\delta= \epsilon \text{sign}(w)$. If $w_i = \mathcal{O}(1)$ then $\Vert w^T\delta \Vert = \mathcal{O}(d\delta )$, so that $f(\tilde{x})-f(x)$ is much larger than $\delta$. This intuition gave rise to the \emph{Fast Gradient Sign Method}, which is a single-step method to get an adversarial example by taking $\epsilon$ step in the opposite direction of the loss gradient, that is, $\delta = \epsilon \text{sign}(\nabla_x \ell(w,x,y))$. Although this method is fast, it is not efficient, which enabled the development of an iterative approach to creating $\tilde{x}$ called \emph{Projected Gradient Descent} (PGD)~\cite{kurakin2016adversarial, madry2018towards}. In PGD, the adversarial example $\tilde{x}$ is obtained by repeated movement along the gradient of loss, projecting back to $\mathcal{B}_{\epsilon}(x)$ if moved out of the budget. Formally, 
\begin{equation}
x^{t+1}= \Pi_{\mathcal{B}_{\epsilon}(x^t)} \lb x^t + \alpha \text{sign}(\nabla_{x^t} \ell(w,x^t,y)) \rb,
\end{equation}
where $\alpha$ is the step size and $\Pi$ denotes the projection operator. For instance, using the $\ell_\infty$ norm, adversarial examples for MNIST can be obtained with $\epsilon=0.3$, $\alpha=0.01$ and $20$ or $40$ PGD steps, and for CIFAR-10 with $\epsilon=\frac{8}{255}$, $\alpha=\frac{2}{255}$ and $40$ PGD steps. It is sufficient to perturb the data with suggested $\epsilon$ and $\alpha$ to completely break down the predictions of trained models.

\paragraph{Adversarial Training.} Since the discovery of adversarial examples \cite{Sze+14}, many approaches have been proposed for training models that are resistant to attacks. The most prominent one, which enjoys both simplicity and empirical success, is called \textit{Adversarial Training} \cite{goodfellow2014explaining,madry2018towards}, where the standard empirical risk minimization procedure is replaced by a worst-case version of it. That is, instead of using the training data to fit the model parameters, we use the worst possible data with respect to the model. Formally, adversarial training solves the following optimization problem:
\begin{equation}\label{eq:at}
    \min_\theta \mathbb{E}_{(x,y)\sim \mathcal{D}} \lb \max_{\delta \in \mathcal{B}_\epsilon(x)} \ell (f_\theta(x+\delta), y) \rb,
\end{equation} where $\ell$ is a classification loss (typically cross entropy). In practice, this problem is being solved by alternating a step of gradient descent on $\theta$ and one of (projected) gradient ascent on $\delta$. Several variations of the above framework have been proposed in the literature. A particularly noteworthy one is called \textit{TRADES} \cite{Zha+19} and lies on the observation that there seems to be a tension between classification accuracy and robustness (see also Section \ref{ssec:tradeoff}). For that reason, the worst case loss in \eqref{eq:at} is being decomposed in two terms; one that captures clean classification accuracy and one that penalizes different outputs inside the allowed perturbation set. Formally the optimization problem becomes
\begin{equation}\label{eq:trades}
    \min_\theta \mathbb{E}_{(x,y)\sim \mathcal{D}} \lb \ell (f_\theta(x), y) + \lambda \max_{\delta \in \mathcal{B}_\epsilon(x)} \ell \left(f_\theta(x+\delta), f_\theta(x)\right) \rb,
\end{equation} 
where $\lambda$ is a parameter that controls the tradeoff between clean and robust accuracy.

However, it is possible to find a kernel formulation called Adversarial KIP~\cite{TSK22} inspired by the data distillation method discussed in the previous section to tackle the problem of robust classification.
\begin{align}
    \Ls_{Adv KIP} &= \Vert Y_T - \mathcal{K}(X_T+\tilde{\delta},X_S)\mathcal{K}(X_S,X_S)^{-1}Y_S \Vert_2, \text{ where } \nonumber \\
    \tilde{\delta}&= \mathrm{arg} \max_{\delta \in \Delta(0)} \ell(\mathcal{K}(X_T+\delta,X_S)\mathcal{K}(X_S,X_S)^{-1}Y_S, Y_T) \nonumber
\end{align}
Here loss function $\ell(.)$ can be different for learning $X_S$ and $\delta$, for example, squared error and cross-entropy losses for learning $X_S$ and $\delta$, respectively.

\subsection{Adversarial robustness requires more data}

In this section, we discuss the sample complexity separation result from \cite{schmidt2018adversarially} which theoretically shows that adversarial training requires more data compared to standard training to attain good generalization. The analysis considers a simple Gaussian model and it is focused on $\ell_\infty$-robustness. The specific setup is described below.
 
\paragraph{Setup.} Let $(x,y) \in \mathbb{R}^d \times \{\pm 1\}$ be the data, $\theta^* \in \mathbb{R}^d$ be the per-class mean vector and $\sigma>0$ be the variance, then $(\theta^*, \sigma)$ is the Gaussian model defined by the distribution $x \sim \mathcal{N}(y \theta^*, \sigma^2 I)$. After fixing the norm of $\theta^*$ to $\sqrt{d}$, the tunable parameter is the variance $\sigma^2$ which controls the amount of overlap between the two classes. As $\ell_\infty$-robustness is considered, the perturbation set is $\mathcal{B}_\infty^\epsilon(x) = \{ x^\prime \in \mathbb{R}^d | \Vert x^\prime - x \Vert_\infty \leq \epsilon\}$. The model used to learn is the linear classifier $f_w(x^\prime) = \text{sign}(w^T x^\prime)$.

\paragraph{Sample complexity for standard training.} Let $(x,y)$ be sampled from the Gaussian model $(\theta^*, \sigma)$ with $\Vert \theta^* \Vert_2 = \sqrt{d}$ and $\sigma \leq c d^{1/4}$ where $c$ is some constant. Then $1$ sample is enough to learn the well-generalizing linear classifier $f_{\hat{w}}$ with $\hat{w}=yx$. The proof idea is as follows: since the Gaussian distribution is rotationally invariant, we are free to use a basis where $\theta^*= (\sqrt{d}, 0, \ldots,0)$. Then 
\begin{align}
    yx &= (\sqrt{d} + \mathcal{N}(0,c^2 d^{1/2}), \mathcal{N}(0,c^2 d^{1/2}), \ldots,  \mathcal{N}(0,c^2 d^{1/2}) ) \nonumber \\
    &\Rightarrow \linnerprod \theta^*, yx \rinnerprod = d + \mathcal{N}(0, c^2d^{1/2}d). \nonumber
\end{align}
The prediction is wrong when $\mathcal{N}(0, c^2d^{1/2}d)$ is negative and larger than $d$ in modulus, therefore the probability of misclassification is controlled by $c^2d^{1/2}d$. It can be restricted by choosing small enough $c$, that is, one can choose a $c$ such that the misclassification error is $<1\%$. For this, the above equation needs to be modified to actually compute the generalization error sign$(wx)$ and define $w$ to be the one sample.

\paragraph{Sample complexity for adversarial training.} Let $(x_i,y_i)_{i=1}^{n}$ be sampled independently from the Gaussian model $(\theta^*, \sigma)$ with $\Vert \theta^* \Vert_2 = \sqrt{d}$ and $\sigma \leq c d^{1/4}$ where $c$ is some constant. Then $n = \Omega(\sqrt{d})$ samples are required for learning the robust classifier $\hat{w}=\frac{1}{n} \sum_{i=1}^n y_ix_i$. The proof idea is as follows: in this setup, the model is not rotationally invariant as $\ell_\infty$ perturbations are allowed. Let $$x = \theta^* + (\mathcal{N}(0,c^2 d^{1/2}), \ldots, \mathcal{N}(0,c^2 d^{1/2})).$$ Therefore,
\begin{align}
    \hat{w} &= \frac{1}{n} \sum_{i=1}^n y_ix_i \nonumber \\
    &= \theta^* + \underbrace{\mathcal{N}\lp0, \frac{c^2d^{1/2}}{n}\mathbf{1}_d\rp}_{\text{noise vector } n} \nonumber
\end{align}
Let $(x_{test},y_{test})$ be the test sample with $x_{test}=x_t + \delta$ where $\delta$ is the $\ell_\infty$ perturbation with $\epsilon$ budget (ie) $\Vert \delta \Vert_\infty \leq \epsilon$ and $y_{test}=1$. So, $\linnerprod \theta^* + n, x_{test} \rinnerprod < 0$ to misclassify $x_{test}$.
\begin{align}
    \linnerprod \theta^* + n, x_{test} \rinnerprod &= \linnerprod \theta^* + n, x_{t} + \delta \rinnerprod \nonumber \\
    &= \linnerprod \theta^*, x_{t} \rinnerprod + \linnerprod n, x_{t}\rinnerprod + \linnerprod \theta^*, \delta \rinnerprod + \linnerprod n,\delta \rinnerprod \nonumber \\
    &\simeq d \pm \frac{c d^{1/4} }{\sqrt{n}}\sqrt{d} \pm \epsilon \sqrt{d} \pm \epsilon d \frac{ cd^{1/4}}{\sqrt{n}} \label{eq:samp_comp_adv}
\end{align}
For $ x_{test}$ to be not misclassified, $\linnerprod \theta^* + n, x_{test} \rinnerprod $ should be $>0$, implying that the last term in \eqref{eq:samp_comp_adv} should be significantly less. This means, $n >> \sqrt{d}$. Thus it requires $n = \Omega(\sqrt{d})$ samples to robustly learn under $\ell_\infty$ perturbation.

\paragraph{}It is clear from the above analysis that adding Gaussian noise doesn't really affect the  performance of the classifier. However, standard trained models are extremely sensitive toward adversarial perturbation. Recent works in the line of geometric analysis, following the empirical evidence that real data lie in a low-dimensional manifold~\cite{pope2021intrinsic}, also provide a similar insight on model robustness. Further research \cite{khoury2018geometry} analyzed the hypothesis that the low-dimension manifold is embedded in a high-dimensional manifold, thus allowing the decision boundaries to be manipulated in different ways by the adversary. There is also recent progress on separation results by  \cite{bubeck2021universal}, in which it is shown that robust models are computationally hard to obtain. For instance, to find a robust classification model that is Lipschitz, training requires a number of parameters inversely proportional to the Lipschitz constant. This is also empirically shown in \cite{madry2018towards}. All these results imply that training a robust model is both a data-hungry process and requires a lot of parameters. 

\subsection{Trade-off between accuracy and robustness}\label{ssec:tradeoff}
There are challenges in training robust classifiers such as more data and larger model complexity. In addition to these computational difficulties, such robust models perform poorly on standard test data. It is shown theoretically and empirically in \cite{tsipras2018robustness}. 

\paragraph{Theoretical intuition.} Let $X = (x_0, \ldots, x_d) \in \mathbb{R}^{d+1}$ and $Y \in \{\pm 1\}$ be the data features and label. The data is constructed as follows: 
\begin{align}
    Y &\sim \text{Unif}(\{\pm 1\}) \nonumber \\
    x_0 &= 
    \begin{cases}
    y & w.p. \quad 0.9 \nonumber \\
    -y & w.p. \quad 0.1
    \end{cases} \nonumber \\
    x_1, \ldots, x_d &\sim \mathcal{N}\lp \frac{10 y}{\sqrt{d}}, 1 \rp \nonumber
\end{align}
The features $x_0$ are strongly correlated to the label $Y$ and $x_1, \ldots, x_d$ are weakly correlated.
The classifier $f_w(X)=\text{sign}(w^TX)$ can use either feature $x_0$ alone or $x_1, \ldots, x_d$ to classify with minimal error. Lets first consider $w = \frac{1}{d}(0, 1,\ldots, 1)$ (ie) weighted average of $x_1, \ldots, x_d$. The probability of the classifier with $w$ predicting the labels correctly is
\begin{align}
    \mathbb{P} \lp \text{sign}(w^TX)=Y \rp &= \mathbb{P}\lp \frac{1}{d}\sum_{i=1}^d \mathcal{N}\lp \frac{10}{\sqrt{d}},1 \rp > 0\rp > 0.99 \nonumber
\end{align}
This implies we get much better performance with the weakly correlated feature than the strongly correlated feature $x_0$. If an adversary is allowed with a small budget $\epsilon = \frac{20}{\sqrt{d}}$, then it can flip the sign of all $x_1, \ldots, x_d$ and completely alter the prediction of the classifier. Hence, the features $x_1, \ldots, x_d$ are useful but non-robust whereas $x_0$ is a robust feature. One can note that relying on the robust feature $x_0$ will lower the performance on standard samples, increasing however the robustness. This simple example highlights the tension that exists between robustness and accuracy. More details can be found in \cite{tsipras2018robustness}. 

The above toy example formalizes the intuition behind several empirical works that propose the existence of \textit{robust} and \textit{non-robust} features \cite{Ily+19} in the data.

\subsection{Robust and Non-Robust Features}

\sloppy
Let $\mathcal{D}$ be a data distribution and sample pairs $(x, y) \in \mathcal{X} \times \{\pm 1 \}$. We define features to be the set, $\mathcal{F}$, of all (measurable) functions from the input space to the reals, i.e. $\mathcal{F} = \{f : \mathcal{X} \to \mathbb{R}\}$. For convenience, also assume that each $f \in \mathcal{F}$ is centered with respect to $\mathcal{D}$, that is $\mathbb{E}_{(x, y) \sim \mathcal{D}} [f(x)] = 0$ and $\mathbb{E}_{(x, y) \sim \mathcal{D}} [f^2(x)] = 1$. \cite{Ily+19} define the following subsets of $\mathcal{F}$:

\begin{itemize}
    \item $\rho$-\textbf{useful features}: We call a feature $f$ $\rho$-\textit{useful} ($\rho > 0$) if, in expectation, it is correlated with the true label:
    \begin{equation}
        \mathbb{E}_{(x, y) \sim \mathcal{D}} \lb yf(x)\rb \geq \rho.
    \end{equation}
    
    \item $\gamma$-\textbf{robustly useful features}: Suppose we have an $f \in \mathcal{F}$ that is $\rho$-useful. We refer to $f$ as $\gamma$-\textit{robustly useful} ($\gamma > 0$) if, under any adversarial perturbation in a set $\Delta$, it remains $\gamma$-useful:
    \begin{equation}
        \mathbb{E}_{(x, y) \sim \mathcal{D}} \lb\inf_{\delta \in \Delta(x)} yf(x+ \delta)\rb \geq \gamma.
    \end{equation}
    
    \item \textbf{Useful, non-robust features}: A useful, non-robust feature is a feature which is $\rho$-useful for some $\rho$ bounded away from zero, but is not a $\gamma$-robust feature for any $\gamma \geq 0$. These features help with classification in the standard setting but may hinder accuracy in the adversarial setting, as the correlation with the label can be flipped (as we saw in the previous subsection).
\end{itemize}

In a set of experiments, the authors of \cite{Ily+19} demonstrate that common computer vision datasets consist of both robust and non-robust features (according to the previous definitions). In particular, it is shown that a dataset can be modified so that it only contains robust features and this is enough for robust classification, without the need for specialized algorithms like adversarial training. The findings of \cite{Ily+19} caused a lot of fruitful discussion around the nature of adversarial examples in machine learning \cite{Eng+19}, and, naturally, invited the question of what are these infamous non-robust features.

An answer can be given through the NTK of a neural network. Recall that in the infinite width limit, a converged neural network is equivalent to a kernel machine $f_\infty$ that uses the NTK of the model
\begin{equation}
    f_\infty(x) = K(x, X_T)^\top K(X_T, X_T)^{-1} Y_T.
\end{equation}
By obtaining the eigendecomposition of the Gram matrix $K(X_T, X_T) = \sum_{i = 1}^{\mid X_T \mid} \lambda_i v_i v_i^\top$, one can write
\begin{equation}
    f_\infty(x) = K(x, X_T)^\top \left( \sum_{i = 1}^{\mid X_T \mid} \lambda_i^{-1} v_i v_i^\top \right) Y_T = \sum_{i = 1}^{\mid X_T \mid} f^{(i)}(x),
\end{equation}
where $f^{(i)}(x) = K(x, X_T)^\top \left( \lambda_i^{-1} v_i v_i^\top \right) Y_T$. In words, the prediction of the kernel machine can be decomposed to different functions $f^{(i)}$ from the input space to $\mathbb{R}$. As per the definitions at the beginning of the subsection, these functions can be viewed as features, and thus their usefulness and robustness can be studied.

\begin{figure}
    \centering
    \includegraphics[scale=0.4]{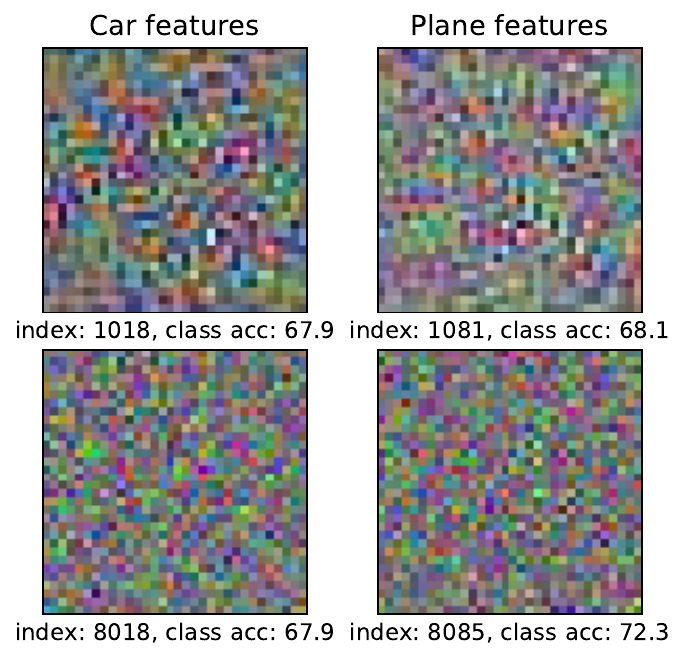}
    \caption{Example of non-robust features that are very accurate in predicting a class on a binary classification task extracted from CIFAR-10 (car vs airplane) \cite{TsKe22}. Index denotes eigenvector index, with the smallest eigenvalue corresponding to 10000.}
    \label{fig:nonrobfeats}
\end{figure}

In \cite{TsKe22}, it was indeed found that for several common architectures and for standard computer vision tasks, the functions $f^{(i)}$ can be split into groups of robust and non-robust ones. Interestingly, this framework allows the visualization of the feature functions, which also reveals that useful, non-robust features appear as random patterns to a human eye---see Figure \ref{fig:nonrobfeats}.


\section{Conclusion}

The analytical tools afforded by the NTK theory have provided algorithms for several problems in machine learning in recent years. Examples include algorithms for automated architecture search \cite{CGW21,Xu+21}, model selection \cite{Des+21}, poisoning attacks \cite{YuWu21} and pruning of large networks\cite{LiZe20,YaWa22}. The formalism of kernels allows one to reduce the complex behavior of a trained neural net to a closed expression that involves directly the architecture and the training data, which is invaluable when one wants to suitably optimize one of the two.

As it was mentioned before, an NTK approach can also be taken in the context of robust classification. Extending the KIP formulation of \cite{Ngu+21}, \cite{TSK22} introduced an algorithm, coined \textit{Adversarial KIP}, for learning a dataset that produces classifiers that are both accurate and robust.

It remains an active research question, however, whether the kernel regime is relevant for neural networks used in practice. To this end, several recent works \cite{Fort+20,Bar+21,Jim+21} approached this problem from an empirical point of view: they deployed standard architectures that operate in a non-lazy fashion (meaning that the weights of the network change significantly during training) and measured how much the NTK quantity \eqref{eq:NTK} deviates from its analytical prediction. The common consensus from these empirical studies is that the NTK undergoes a phase of rapid evolution at the beginning of training, followed by a period when the kernel stabilizes and changes only in scale (long before the convergence of the network, in terms of loss). Interestingly, a similar behavior was observed during adversarial training \cite{TsKe22}. What these works seem to suggest is that only a few epochs of training suffice to provide kernel quantities that describe accurately the behavior of the final network, and perhaps these would be better options for designing practical machine learning algorithms (instead of using the infinite-width quantities) in the future.

\paragraph{Acknowledgements}
These are notes from the lecture of Julia Kempe given at the summer school "Statistical Physics \& Machine Learning", that took place in Les Houches School of Physics in France from 4th to 29th July 2022. The school was organized by Florent Krzakala and Lenka Zdeborová from EPFL. For the author D. O. this work was partially supported by FAPERJ (E-26/202.668/2019) and CAPES (Brazil).

\printbibliography

\end{document}